\documentclass[10pt,twocolumn,letterpaper]{article}
\usepackage{lineno}
\usepackage{cvpr}              % To produce the CAMERA-READY version
\usepackage{multirow}
\usepackage{colortbl}
\usepackage{xcolor}

\usepackage{graphicx}
\usepackage{color}
\usepackage{overpic}

\definecolor{cvprblue}{rgb}{0.21,0.49,0.74}
\usepackage[pagebackref,breaklinks,colorlinks,allcolors=cvprblue]{hyperref}

\def\paperID{1101} % *** Enter the Paper ID here
\def\confName{CVPR}
\def\confYear{2025}

\title{LITA-GS: Illumination-Agnostic Novel View Synthesis via Reference-Free 3D Gaussian Splatting and Physical Priors}

\author{Han Zhou \quad Wei Dong$^\dagger$ \quad Jun Chen\\
McMaster University \quad $^\dagger$ Corresponding Author\\
{\tt\small \{zhouh115, dongw22, chenjun\}@mcmaster.ca}
}

\begin{document}
%\maketitle
%
{
\twocolumn[{
\renewcommand\twocolumn[1][]{#1}
\maketitle

\begin{center}
\setlength{\abovecaptionskip}{0.7mm}
    \includegraphics[width = 0.96\textwidth]{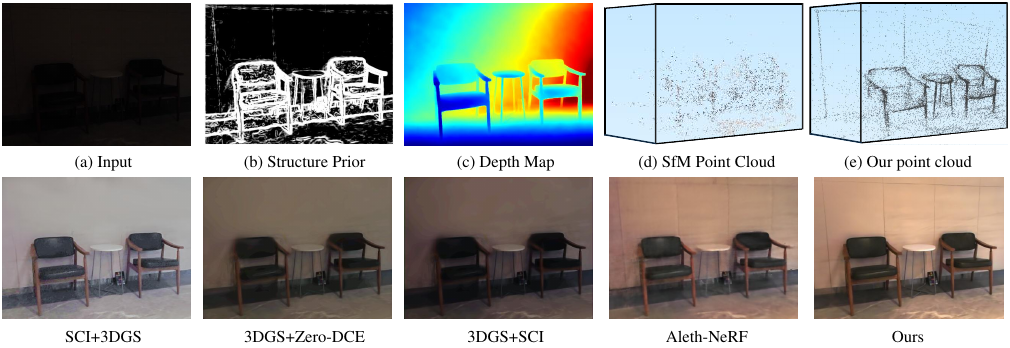}
    \captionof{figure}{First row: (a) Input image; (b) Our rendered illumination-invariant structure prior; (c) Our rendered depth map; (d) SfM point cloud estimated from low-light scenes; (e) Our optimized point cloud. The second row provides visual comparisons between our method and other SOTA approaches. With the rendering of illumination-invariant structure prior and depth map, our method effectively represents the structure and spatial geometry of the scene, thereby achieving superior performance compared to current SOTA approaches.}
    \vspace{1mm}
    \label{fig1}
\end{center}}]
}

\begin{abstract}
 Directly employing 3D Gaussian Splatting (3DGS) on images with adverse illumination conditions exhibits considerable difficulty in achieving high-quality, normally-exposed representations due to: (1) The limited Structure from Motion (SfM) points estimated in adverse illumination scenarios fail to capture sufficient scene details; (2) Without ground-truth references, the intensive information loss, significant noise, and color distortion pose substantial challenges for 3DGS to produce high-quality results; (3) Combining existing exposure correction methods with 3DGS does not achieve satisfactory performance due to their individual enhancement processes, which lead to the illumination inconsistency between enhanced images from different viewpoints. To address these issues, we propose \textbf{LITA-GS}, a novel illumination-agnostic novel view synthesis method via reference-free 3DGS and physical priors. Firstly, we introduce an illumination-invariant physical prior extraction pipeline. Secondly, based on the extracted robust spatial structure prior, we develop the lighting-agnostic structure rendering strategy, which facilitates the optimization of the scene structure and object appearance. Moreover, a progressive denoising module is introduced to effectively mitigate the noise within the light-invariant representation. We adopt the unsupervised strategy for the training of LITA-GS and extensive experiments demonstrate that LITA-GS surpasses the state-of-the-art (SOTA) NeRF-based method while enjoying faster inference speed and costing reduced training time. The code is released at \url{https://github.com/LowLevelAI/LITA-GS}.
\end{abstract}

%by 1.7 dB in PSNR and 0.09 in SSIM    
\section{Introduction}
\label{sec:intro}
Novel view synthesis is an important task in computer vision and has wide applications in augmented and virtual reality (AR/VR). The advent of Neural Radiance Fields (NeRF)~\cite{nerf} and 3D Gaussian Splatting (3DGS)~\cite{3Dgaussian} has led to unprecedented progress and achievements in this field. For example, existing methods are capable of delivering high-quality novel views and offering real-time rendering and accelerated training. Yet, it is imperative to acknowledge that these considerable accomplishments rely on having multiple well-exposed images as a preliminary condition. In real-world scenarios such as over-exposed urban surveillance, nighttime driving, and robotic exploration and rescue operations in dark environments, the majority of existing novel image synthesis methods fail to perform adequately. This deficiency underscores the necessity of developing additional modules specifically engineered to analyze and correct the adverse lighting conditions. 

A number of NeRF-based methods have endeavored to tackle the difficulties associated with adverse lighting conditions. Specifically, each 3D point in LLNeRF~\cite{LightingupNeRF} is decomposed into a view-independent basis component and a light-related view-dependent component. These components are manipulated to enhance the brightness, correct the colors and reduce the noise. AlethNeRF~\cite{alethnerf} introduces the concept of concealing field to interpret the lightness degradation, and such concealing field is employed or removed to achieve normal-light rendering under over-exposed or low-light conditions.
However, like all NeRF-based novel synthesis methods, these techniques share a common drawback: the prohibitively long training times and the inability to achieve real-time rendering. This limitation restricts their practical applications and underscores the urgent demand for novel image synthesis technologies that can support real-time rendering while effectively handling adverse illumination conditions.

The new emergent 3DGS~\cite{3Dgaussian} has demonstrated impressive capability in producing high-quality novel images and offering real-time rendering speed by employing a set of 3D Gaussian primitives to reconstruct the scene. However, it is infeasible to directly train the vanilla 3DGS using images captured under environments with adverse illumination. First, the performance on 3DGS heavily depends on the quality of Structure from Motion (SfM)~\cite{colmap} points, while the limited SfM points estimated in adverse illumination scenarios, especially in dark environments, fail to capture sufficient scene details. Second, under-exposed and over-exposed images presents substantial information loss, intensive noise, and severe color distortion, which pose great challenges for 3DGS to produce high-quality normally-exposed novel images, especially when no ground truth images are available during the training. Besides, introducing current exposure correction or image restoration methods \cite{zhou2024glare, zhou2024gppllie, dong2024ecmamba, zhou2023breaking, dong2024dehazedct, dong2024shadowrefiner} as the pre-/post-precessing tool for 3DGS also brings new problem: these approaches primarily focus on the enhancing the images in their original views individually rather than generating coherent 3D scenes, which leads to the illumination inconsistency between enhanced images from different viewpoints.

To tackle these issues, we propose an illumination-agnostic novel view synthesis via reference-free 3D
Gaussian Splatting and physical priors (denoted as \textbf{LITA-GS}) in this paper. Given the challenges in representing structures and details through limited SfM points extracted from images with abnormal exposure, we firstly introduce an illumination-invariant physical prior extraction pipeline. Then, we design the lighting-agnostic structure rendering based on the extracted robust spatial structure prior, which facilitates the optimization of 3D Gaussians. Moreover, we render both illumination and noise components within our lighting-agnostic structure rendering process. The illumination component is instrumental in decomposing the illumination-invariant basis components from those related to light, whereas the noise component is processed through our progressive denoising module for effective noise suppression.
We essentially train our LITA-GS without GT references and our LITA-GS achieves superior performance and enjoys faster convergence and rendering speed than current SOTA methods.

Our contribution can be summarized as follows:

\begin{enumerate}
    \item [1.] We introduce LITA-GS, the first 3DGS-based unsupervised framework for producing high-quality 3D scene representation under adverse illumination conditions.
    \item [2.] To enhance the scene structure, we develop lighting-agnostic structure rendering based on the spatial structure prior extracted by our introduced illumination-invariant physical prior extraction pipeline. 
    \item [3.] Moreover, a lightweight progressive denoising module is proposed based on noise rendering to suppress the noise.
    \item [4.] Extensive experiments demonstrate that our LITA-GS significantly outperforms SOTA NeRF-based methods with much faster speed. Compared to combining exposure correction methods with 3DGS, our LITA-GS achieves superior performance with improved multi-view consistency.
\end{enumerate}
\begin{figure*}[ht]
    \setlength{\abovecaptionskip}{2mm}
    \centering
    \begin{overpic}[width=1\textwidth]{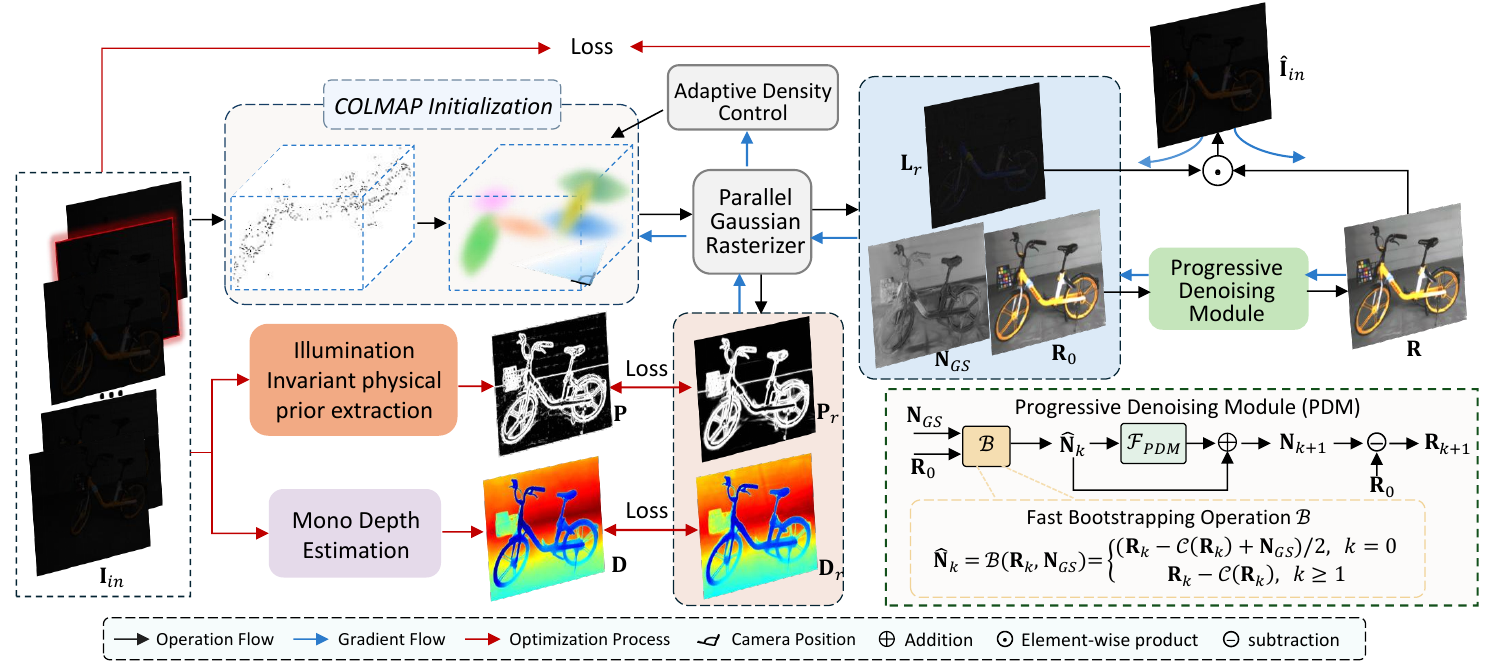}
    \end{overpic}
    \caption{The framework of our \textbf{LITA-GS}. We introduce a physical prior extraction pipeline to capture structural information $\mathbf{P}$ from images with low or high exposure. Then, the extracted $\mathbf{P}$ is integrated into our developed lighting-agnostic structure rendering process. Furthermore, we employ a progressive denoising module (PDM) for noise reduction and optimize our LITA-GS without GT references.}
    %\vspace{-2mm}
    \label{fig_model}
\end{figure*}

\section{Related Works}
\label{sec:rela}
\paragraph{3D Scene Representation for NVS in adverse illumination conditions:} 
NeRF~\cite{nerf} has gained popularity for its ability to generate photorealistic 3D views from limited data using deep neural networks. Subsequent works have extended NeRF for 3D reconstruction of scenes with challenging lighting conditions. 
NeRF-W~\cite{Nerf-w} addresses variable lighting and transient occlusions in unstructured image collections by incorporating image-dependent radiance adjustments and identifying and managing transient elements within scenes.
RawNeRF~\cite{rawnerf} proposes training NeRF directly on RAW data can effectively handle noise in dark scenes.
Given a set of commonly used sRGB images captured in low-light scenes, LLNeRF~\cite{LightingupNeRF} decomposes the color of 3D points into illumination-related view-dependent and view-independent components during NeRF optimization, facilitating the enhancement of novel view images. Aleth-NeRF~\cite{alethnerf} integrates the concealing field assumption into NeRF to adapt to varying illumination conditions.  

However, the ray-tracing regime and the employment of the Multi-Layer Perceptron generally results in slow training and inference speeds for these NeRF-based methods.

Recently, 3D Gaussian Splatting (3DGS)~\cite{3Dgaussian} excels by rapidly rendering complex scenes through efficient rasterization and blending of 3D Gaussian primitives. Nevertheless, the potential for using 3DGS in scene reconstruction under varying lighting conditions, such as low light and overexposure, remains unexplored. Moreover, despite the challenges in obtaining Ground-Truth data, most current 3DGS methods continue to rely on supervised training, underscoring the need for further investigation into unsupervised parameter optimization techniques.

\paragraph{Initialization in Gaussian Splatting}
3DGS is typically initialized using sparse points from Structure-from-Motion (SfM)~\cite{colmap} or random generation. However, as shown in ~\cite{relaxing}, a noisy, inaccurate SfM-initialized point cloud can trap 3DGS in local minima, reducing performance. To circumvent this, DUSt3R~\cite{dust3r} employs a Siamese architecture composed of a shared ViT encoder to obtain a pointmap. Later methods~\cite{gaussianobject, LMGaussian, instantsplat} have also adopted this COLMAP-free approach.
Despite achieving promising performance, this method of initializing point clouds is time-consuming, thereby impairing the training speed and limiting its suitability for scenarios demanding rapid processing and real-time performance.

\paragraph{Improving Gaussian Splatting with Extra Attributes} Some recent studies aim to enhance the rendering capabilities of 3DGS by adding new attributes. For instance, ~\cite{mirrorgaussian} derives the mirrored counterpart of the real-world scene by incorporating a mirror label and a mirror plane attribute.
~\cite{Urban3D,feature,street} introduce semantic attributes to enable the model to understand complex scenes. ~\cite{gaussian_grouping} incorporates Identity Encoding within 3D Gaussians, grouping them by object instances to facilitate versatile scene editing tasks. 
~\cite{Uncertainty} embeds language features and learned uncertainty values into 3D Guassians to mitigate semantic ambiguities arising from visual inconsistency in multi-view images.

\section{Preliminary}
\label{sec_pre}
3D Gaussian Splatting (3DGS) represents a 3D scene using $K$ anisotropic 3D Gaussian primitives, $\{G_i | i=1,...,K \}$, where each Gaussian $G_i$ is parameterized by an opacity (scale) $\alpha_i\in[0,1]$, a center $\mu_i \in \mathbb{R}^{3}$ and a covariance matrix $\Sigma_i \in \mathbb{R}^{3 \times 3}$ defined in the world space: 
\begin{equation}
	G_i(x)~= e^{-\frac{1}{2}(x-\mu_i)^T \Sigma_i^{-1}(x-\mu_i)},
    \label{3DGS} 
\end{equation}
where $\Sigma_i$ is defined by a rotation matrix $R_i \in \mathbb{R}^{3 \times 3}$ and a scaling matrix $S_i \in \mathbb{R}^{3 \times 3}$, as $\Sigma_i = R_i S_i S^T_i R^T_i$, ensuring positive semi-definiteness. For separate optimization, $R_i$ and $S_i$ are stored as a rotation quaternion $q_i\in \mathbb{R}^{4}$ and a scaling factor $s_i\in \mathbb{R}^{3}$, respectively.

Besides, for each 3D Gaussian, spherical harmonic coefficients are utilized to model view-dependent color $c_i$.

\section{Method}
\label{sec_method}
 In this work, to achieve satisfactory scene representation for real-world scenarios with adverse illumination conditions, we develop a novel method named \textbf{LITA-GS} for illumination-agnostic novel view synthesis via reference-free learning. In this section, we firstly introduce our proposed illumination-invariant prior extraction pipeline (Sec.~\ref{sec_prior}). Secondly, we detail the lighting-agnostic spatial structure rendering, where extra attributes (\textit{i.e.,} illumination-invariant structure, illumination feature and noise representation) are attached for each Gaussian (Sec.~\ref{sec_rendering}). Then, we design a progressive denoising module for noise suppression (Sec.~\ref{sec_denosing}). The overall framework of the proposed \textbf{LITA-GS} is illustrated in Fig.~\ref{fig_model}, and the loss function of our reference-free optimization process is provided in Sec.~\ref{sec_loss}.

\subsection{Illumination-Invariant Prior Extraction}
\label{sec_prior}
Our illumination invariant physical prior is extracted based on Kubelka-Munk theory~\cite{Kubelka-Munk-theory, Kubelka-Munk-theory2, Kubelka-Munk} and the invariant edge detectors~\cite{edge, colorinvariance}. Specifically, the reflected light energy $E$ for an object in the image space is calculated by:
\begin{equation}
    E(\lambda,\mathbf{z}) = e(\lambda,\mathbf{z})\left((1-r_f(\mathbf{z}))^2R_\infty(\lambda,\mathbf{z})+r_f(\mathbf{z})\right),
    \label{eq_energy}
\end{equation}
where $e(\lambda,\mathbf{z})$ denotes the illumination spectrum, $\mathbf{z}=(x,y)$ represents position at the imaging plane, $\lambda$ denotes the wavelength of the light, $r_f(\mathbf{z})$ the Fresnel reflectance at $\mathbf{z}$, and $R_\infty(\lambda,\mathbf{z})$ is the material reflectivity. For matte and dull surfaces, the Fresnel coefficient is generally negligible, $r_f(\mathbf{z})\approx 0$ and the Eq.~\ref{eq_energy} can be simplified as :
\begin{equation}
    E(\lambda,\mathbf{z}) = e(\lambda,\mathbf{z})\left(R_\infty(\lambda,\mathbf{z})\right).
    \label{eq_energy_mette}
\end{equation}

Assuming equal energy and uniform illumination, the $e(\lambda,\mathbf{z})$ in Eq.~\ref{eq_energy_mette} can be regarded as a constant $i$, then the differentiation of $E$ with respect to $\mathbf{z}$, denoted as $\nabla_{\mathbf{z}}E$ and the ratio $\nabla_{\mathbf{z}}P=\frac{\nabla_{\mathbf{z}}E}{E}$ are as follows:
\begin{equation}
\begin{aligned}
    \nabla_{\mathbf{z}}E &= \frac{\partial E}{\partial \mathbf{z}} = i\frac{\partial R_\infty}{\partial \mathbf{z}}, \quad
    \nabla_{\mathbf{z}}P = \frac{1}{ R_\infty}\frac{\partial R_\infty}{\partial \mathbf{z}},
\label{eq_illu_invariant}
\end{aligned}   
\end{equation}
%\nabla_{\mathbf{z}}E
where $\nabla_{\mathbf{z}}P$ quantifies variations in object reflectance independently of the illumination intensity. The same holds for the ratios $\nabla_{\lambda\mathbf{z}}P=\frac{\nabla_{\lambda\mathbf{z}}E}{E}$ and $\nabla_{\lambda\lambda\mathbf{z}}P=\frac{\nabla_{\lambda\lambda\mathbf{z}}E}{E}$, where $\nabla_{\lambda\mathbf{z}}E$ and $\nabla_{\lambda\lambda\mathbf{z}}E$ can be interpreted respectively as the spatial derivatives of the spectral slope and the spectral curvature. 

Consequently, the illumination invariant edge detector $\mathbf{P}$ can be defined by the gradient magnitude of relevant spatial derivatives as follows:
\begin{equation}
    \mathbf{P} = \sqrt{(\nabla_{\mathbf{z}}P)^2 + \beta(\nabla_{\lambda\mathbf{z}}P)^2 + \gamma(\nabla_{\lambda\lambda\mathbf{z}}P)^2},
\label{eq_final_illumintion_prior}
\end{equation}
where $\beta$ and $\gamma$ are two coefficients to  balance each illumination invariant, and they are set to 1.0 in this paper. Note that we have omitted $(\lambda,\mathbf{z})$ from $E(\lambda,\mathbf{z})$ for simplicity.

Moreover, the spatial derivative $\nabla_{\mathbf{z}}E$ in Eq.~\ref{eq_illu_invariant} is derived along both the x- and y-directions, denoted as $\nabla_{x}E$ and $\nabla_{y}E$, such that the gradient magnitude is $|\nabla_{\mathbf{z}}E| = \sqrt{(\nabla_{x}E)^2 + (\nabla_{y}E)^2}$. 

According to ~\cite{colorinvariance, Kubelka-Munk-theory2}, well-posed spatial differentiation can be derived from the Gaussian color model. Eq.~\ref{eq_gaussian_color_model} provides a direct transformation matrix from RGB camera sensitivities to estimate  $E(\mathbf{z})$, $\nabla_{\lambda} E(\mathbf{z})$, and $\nabla_{\lambda\lambda} E(\mathbf{z})$. Spatial derivatives are then obtained through convolution with a Gaussian derivative kernel $f$, as detailed in Eq.~\ref{eq_gaussian_derivate}.
\begin{equation}
\begin{bmatrix}
  E(\mathbf{z})\\
  \nabla_{\lambda} E(\mathbf{z})\\
  \nabla_{\lambda\lambda} E(\mathbf{z})\\
  \end{bmatrix}
  =
  \begin{bmatrix}
  0.06 & 0.63 & 0.27 \\
  0.3 & 0.04 & -0.35 \\
  0.34 & -0.6 & 0.17 \\
  \end{bmatrix}
  \begin{bmatrix}
  R(\mathbf{z})\\G(\mathbf{z})\\B(\mathbf{z})
  \end{bmatrix}
\label{eq_gaussian_color_model}
\end{equation}
\begin{equation}
\nabla_{x}E(x,y) = \sum_{s \in \mathbb{Z}} E(s,y) \frac{\partial f(x-s,\sigma)}{\partial x},
\label{eq_gaussian_derivate} 
\end{equation}
where $\sigma$ is a hyperparameter that denotes the standard deviation of $f$, and $s \in \mathbb{Z}$ indicates that the summation encompasses all $x$-values within the image space. 

\begin{figure}[!h]
    \setlength{\abovecaptionskip}{2mm}
    \centering
	% 11111111111
        \begin{minipage}[b]{1\linewidth}
		  \centering
		  \includegraphics[width=\linewidth]{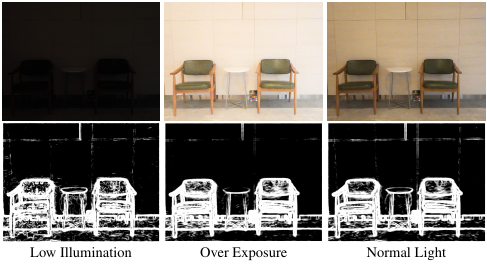}
	\end{minipage}
	\caption{Our extracted $\mathbf{P}$ is invariant to varying illumination.}
	\label{fig_prior_compare}
 \vspace{-6mm}
\end{figure}

\paragraph{Physical Explanation} Eq.~\ref{eq_energy}-\ref{eq_gaussian_derivate} mathematically indicates that $\mathbf{P}$ characterizes the spatial derivatives of spectral intensity. We visualize our extracted structure prior $\mathbf{P}$ for images with different illumination conditions in Fig.~\ref{fig_prior_compare}. It is clear that $\mathbf{P}$ is capable of producing a more stable edge and structure map across images with differing exposure levels, highlighting its great potential to assist 3DGS in effective 3D scene reconstruction under challenging illumination.

\subsection{Lighting-Agnostic Spatial Structure Rendering}
\label{sec_rendering}

Under adverse illumination conditions, Structure from Motion (SfM)~\cite{colmap} algorithms often struggle to predict camera poses and point clouds, as presented by the cloumn 4 in Fig.~\ref{fig1}. Such a sparse point cloud, which fails to effectively convey the scene’s structure, presents difficulties in obtaining high-quality results through vanilla 3DGS, particularly when employing a reference-free optimization strategy. In contrast to supervised learning framework where GT references is utilized as the guidance, the relatively relaxed constraints in reference-free approach pose challenges for 3DGS in accurately locating Gaussian primitives, thus resulting in smoothed appearance and suboptimal background representation. Motivated by the robust edge and structural representations extracted by our  pipeline proposed in Sec.~\ref{sec_prior}, we propose to employ this lighting-invariant structure to enhance the optimization of 3DGS. 

As with many 3DGS-based approaches, our method begins with Gaussian initialization based on SfM estimation. Given the estimated point cloud, the centers of the 3D Gaussians are initialized at each point $m_k\in \mathbb{R}^{3}$, where $k$ denotes the number of points in the cloud. To decouple structural information from challenging lighting conditions, we embed an additional learnable attribute $p_i$ in each 3D Gaussian to represent the lighting-independent spatial structure. 

Upon projecting the 3D Gaussians from onto the 2D plane ~\cite{EWA} for a given viewpoint, besides obtaining the enhanced image $\mathbf{R}_0$, we also acquire the corresponding structure map $\mathbf{P}_r$. Similar to the rendering of $\mathbf{R}_0$, each pixel $g$ in $\mathbf{P}_r$ can be computed by performing volume rendering in front-to-back depth order~\cite{order}:
\begin{equation}
\begin{split}
    &\mathbf{P}_r = \sum_{i \in \mathcal{N}} p_i \alpha_i {\mathcal{G}^{2D}_i}(g)\prod_{j=1}^{i-1}(1-\alpha_j {\mathcal{G}^{2D}_j}(g)),\\
    &\mathcal{G}^{2D}_i(g) = e^{-\frac{1}{2}(g-\mu'_i)^T (\Sigma^{2D}_i)^{-1}(g-\mu'_i)},\\
    &\Sigma^{2D}_i = JW \Sigma_i W^{T} J^{T},
\end{split}
\label{eq_render_our}
\end{equation}
where $\mathcal{N}$ is the set of ordered 2D Gaussians overlapping the pixel, $J$ is the Jacobian of the affine approximation of the projective transformation, and $W$ is the world-to-camera transformation matrix. Throughout the training process, the $\mathbf{P}_r$ for an arbitrary viewpoint is optimized to approximate the corresponding illumination-invariant spatial structure $\mathbf{P}$, thereby enabling our method to accurately capture the scene's geometry. 

Moreover, to further improve scene geometry, particularly in terms of depth consistency, we incorporate depth information $\mathbf{D}$ estimated by the monocular depth estimation network Marigold~\cite{Marigold} into the optimization process of the Gaussians. Similar to the integration of illumination-invariant structure prior, one new attribute $d_i$ is attached to each Gaussian primitive and the rendered depth map $\mathbf{D}_r$ (produced by replacing the $p_i$ with $d_i$ in Eq.~\ref{eq_render_our}) is optimized to closely align with $\mathbf{D}$.

\subsection{Progressive Denoising Module}
\label{sec_denosing}
In this section, we aim to develop a module to further suppress the noise and enhance the scene. We attribute the noise present in the rendered image to two aspects: (1) The intensive noise inherent in under-/over-exposed images; (2) Our proposed 3D scene representation is directly built on images with adverse illumination conditions, thus the noise in these low-quality images, especially in dark images, is inevitably enlarged by the exposure correction process. In light of this, we propose to model the noise representation by assigning a noise attribute to each Gaussian primitive. Consequently, the noise map $\mathbf{N}_{GS}$ specific to a given viewpoint can be derived utilizing the similar rendering strategy outlined in Eq.~\ref{eq_render_our}. Moreover, based on the rendered normal-light image $\mathbf{R}_0$ and the noise map $\mathbf{N}_{GS}$, we develop a reference-free progressive denoising module (PDM), which consists of three $3\times3$ convolutions connected by ReLU. 

Specifically, at \(k\)-th stage of PDM, we first develop the following fast bootstrapping operation $\mathcal{B}$ to estimate an initial noise map $\hat{\mathbf{N}}_k$ by: 
\begin{equation}
\hat{\mathbf{N}}_k = 
\begin{cases} 
    (\mathbf{R}_k - \mathcal{C}(\mathbf{R}_k) + \mathbf{N}_{GS}) / 2, &k=0\\
    \mathbf{R}_k - \mathcal{C}(\mathbf{R}_k), & k\geq 1
\end{cases}
\label{eq_pdm_1}
\end{equation}
where $\mathbf{R}_k - \mathcal{C}(\mathbf{R}_k) $ represents the high-frequency noise perceived by the Gaussian filter $\mathcal{C}$. Then, a simple network $\mathcal{F}_{IDM}$ is employed to estimate the refined noise map $\mathbf{N}_{k+1}$ and the denoised image $\mathbf{R}_{k+1}$ by:
\begin{equation}
\begin{split}
    \mathbf{N}_{k+1} &= \hat{\mathbf{N}}_k - \mathcal{F}_{PDM}(\hat{\mathbf{N}}_k), \\
    \mathbf{R}_{k+1} &= \mathbf{R}_0 - \mathbf{N}_{k+1}.
\end{split}
\label{eq_pdm_2}
\end{equation}
In this paper, the PDM is configured with three stages. The bootstrapping operation $\mathcal{B}$ employed in each stage, progressively takes the output from previous stage as input, thereby inherently providing a bridging mechanism and facilitating the convergence of $\mathcal{F}_{PDM}$ across stages.

\begin{table*}
\centering
\setlength{\abovecaptionskip}{2mm}
\renewcommand\arraystretch{1.3}
%\scriptsize
\setlength{\tabcolsep}{1.2pt}
\resizebox{\textwidth}{!}{
\begin{tabular}{c|ccc|ccc|ccc|ccc|ccc|ccc}
\hline
\specialrule{0.6pt}{0.5pt}{0.5pt} 
Scenes & \multicolumn{3}{c|}{``\textbf{\textit{buu}}"} & \multicolumn{3}{c|}{``\textbf{\textit{chair}}"} & \multicolumn{3}{c|}{``\textbf{\textit{sofa}}"} & \multicolumn{3}{c|}{``\textbf{\textit{bike}}"} & \multicolumn{3}{c|}{``\textbf{\textit{shrub}}"} & \multicolumn{3}{c}{\textbf{\textit{mean}}} \\ 
Method \textbar~Metrics & PSNR$\uparrow$  & SSIM$\uparrow$  & LPIPS$\downarrow$  & PSNR$\uparrow$  & SSIM$\uparrow$  & LPIPS$\downarrow$ & PSNR$\uparrow$  & SSIM$\uparrow$  & LPIPS$\downarrow$ & PSNR$\uparrow$  & SSIM$\uparrow$  & LPIPS$\downarrow$ & PSNR$\uparrow$  & SSIM$\uparrow$  & LPIPS$\downarrow$ & PSNR$\uparrow$  & SSIM$\uparrow$  & LPIPS$\downarrow$\\ \specialrule{0.6pt}{0.5pt}{0.5pt}
\hline 
NeRF & 7.51 & 0.291 & 0.448  & 6.04 & 0.147 & 0.594 & 6.28 & 0.210 & 0.568 & 6.35 & 0.072 & 0.623 & 8.03 & 0.031 & 0.680 & 6.84 & 0.150 &0.582 \\
Vanilla 3DGS &7.74  &0.292  &0.459   &6.26  &0.146  &0.761  &6.21  &0.201  &0.918  &6.38  &0.071  &0.822  &8.74  &0.039  &0.604  &7.07  &0.150  &0.713\\ \hline

\multicolumn{19}{c}{NeRF / 3DGS + Image Enhancement Methods} \\ \hline
NeRF + Zero-DCE & 17.81 & 0.833 & 0.357& 12.44 & 0.684 & 0.547 & 14.43 & 0.787 & 0.539 & 10.16 & 0.468 & 0.557 & 12.58 & 0.282 & 0.540 & 13.48 & 0.610 &0.488\\
NeRF + SCI & 7.84 & 0.660 & 0.562 & 12.07 & 0.699 & 0.584 & 10.25 &0.737  &0.626  & \colorbox{yellow!36}{18.84} & 0.637 & 0.565 & 12.38 & 0.358 & 0.587 & 12.27 & 0.618 & 0.585 \\ 
3DGS + Zero-DCE &\colorbox{yellow!36}{18.86}  &\colorbox{yellow!36}{0.890}  &0.191   &13.24  &0.731  &\colorbox{yellow!36}{0.349}  &14.23  &0.767  &{0.586}  &10.56  &0.498  &0.500  &13.26  &{0.430}  &0.272  &14.03  &0.663  &0.380\\
3DGS + SCI &18.33  &0.869  &\colorbox{yellow!36}{0.184}   &11.51  &0.631  &0.406  &12.98  &0.709  &{0.603}  &8.93  &0.364  &0.554  &12.63  &{0.382}  &0.277  &12.88  &0.591  &0.405 \\   \hline

\multicolumn{19}{c}{Image Enhancement Methods + NeRF / 3DGS} \\ \hline
Zero-DCE + NeRF &17.90  & 0.858 & 0.376  &12.58  & 0.721 & 0.460 &\colorbox{yellow!36}{14.45}  & \colorbox{yellow!36}{0.831} & 0.419 & 10.39 & 0.518 & 0.464 & 12.32 & 0.308 & 0.481 & 13.53 & 0.649 &0.432\\ 
SCI + NeRF & 7.76 & 0.692 & 0.525  & 19.77 & 0.802 & 0.674 & 10.08 & 0.772 & 0.520 & 13.44 & 0.658 & 0.435 & 18.16 & 0.503 & 0.475 & 13.84 & 0.689 & 0.510\\   
Zero-DCE + 3DGS &17.92  &\colorbox{orange!30}{0.896}  &\colorbox{orange!30}{0.179}   &12.94  &0.756  &\colorbox{orange!30}{0.303}  &14.42  &\colorbox{yellow!36}{0.831}  &\colorbox{yellow!36}{0.356}  &10.54  &0.539  &\colorbox{yellow!36}{0.401}  &13.10  &0.467  &\colorbox{yellow!36}{0.229}  &13.78  &0.698  &\colorbox{orange!30}{0.294}\\
SCI + 3DGS &7.95  &0.695  &0.501   &\colorbox{orange!30}{21.77}  &\colorbox{orange!30}{0.866}  &0.350  &9.99  &0.750  &0.452  &13.67  &\colorbox{yellow!36}{0.677}  &\colorbox{orange!30}{0.324}   &\colorbox{orange!30}{18.67}  &\colorbox{yellow!36}{0.657}  &\colorbox{red!26}{0.153} &\colorbox{yellow!36}{14.41}  &\colorbox{yellow!36}{0.729}  &\colorbox{yellow!36}{0.356} \\   \hline

\multicolumn{19}{c}{ End-to-end Methods} \\ \hline
Aleth-NeRF &\colorbox{orange!30}{20.22} & 0.859 & 0.315 & \colorbox{yellow!36}{20.93} & \colorbox{yellow!36}{0.818} & 0.468 & \colorbox{orange!30}{19.52} & \colorbox{orange!30}{0.857} & \colorbox{orange!30}{0.354} & \colorbox{orange!30}{20.46} & \colorbox{orange!30}{0.727} & 0.499 & \colorbox{yellow!36}{18.24} & \colorbox{yellow!36}{0.511} & 0.448 & \colorbox{orange!30}{19.87} & \colorbox{orange!30}{0.754} & 0.417\\

\textbf{Ours} &\colorbox{red!26}{20.59}  &\colorbox{red!26}{0.897}  &\colorbox{red!26}{0.175}   &\colorbox{red!26}{22.60}  &\colorbox{red!26}{0.873}  &\colorbox{red!26}{0.223}  &\colorbox{red!26}{20.43} &\colorbox{red!26}{0.895}  &\colorbox{red!26}{0.268}   &\colorbox{red!26}{22.75}  &\colorbox{red!26}{0.819}  &\colorbox{red!26}{0.282} &\colorbox{red!26}{19.35}  &\colorbox{red!26}{0.659}  &\colorbox{orange!30}{0.217}  &\colorbox{red!26}{21.14}  &\colorbox{red!26}{0.829}  &\colorbox{red!26}{0.233}\\ 
\specialrule{0.6pt}{0.5pt}{0.5pt}
\hline
\end{tabular}}
%\vspace{1pt}
\caption{\textbf{Comparison on low-light scenes.} We report PSNR, SSIM, LPIPS and color each cell as \colorbox{red!26}{best}, \colorbox{orange!30}{second best} and \colorbox{yellow!36}{third best}. %Our method has achieved the best rendering quality, while striking a good balance between FPS and the storage memory.
}
\label{tab:1}
 \vspace{-6mm}
\end{table*}

%\cellcolor{yellow!40} third
%\cellcolor{orange!40} second
%\cellcolor{red!40} best

\subsection{Unsupervised Optimization Strategy}
\label{sec_loss}
To enhance the applicability of our method in real-world applications, we implement an unsupervised training strategy to optimize the multiple attributes of Gaussians and the $\mathcal{F}_{PDM}$ network. 

\paragraph{Exposure Control Loss} To facilitate high-quality novel image synthesis, for an arbitrary viewpoint, we employ the exposure control loss $\mathcal{L}_{exp}$ to optimize the rendered image:
\begin{equation}
\begin{split}
&\mathcal{L}_{exp} = \mathcal{L}_1(\mathbf{R}, \hat{\mathbf{I}}_{in}), \\
&\hat{\mathbf{I}}_{in} = \theta / \texttt{mean}(\mathbf{I}_{in}) * \mathbf{I}_{in},
\end{split}
\label{eq_loss_enh}
\end{equation}
where $\mathbf{I}_{in}$ and $\mathbf{R}$ denote the original input image and the final result of $\mathcal{F}_{IDM}$ respectively. Besides, $\texttt{mean}(\mathbf{I}_{in})$ represents the average intensity of $\mathbf{I}_{in}$, and $\theta$ is utilized to generate the modulated image $\hat{\mathbf{I}}_{in}$ with specified intensity degree and enriched structures, thereby facilitating the optimization of illumination-invariant rendered image $\mathbf{R}$. 

\paragraph{Structural Consistency Loss} To regulate the rendered spatial structure prior and depth, we also propose to maximize the structure similarity between these rendered maps ($\mathbf{P}_r$ and $\mathbf{D}_r$) to their corresponding targets ($\mathbf{P}$ and $\mathbf{D}$). Specifically, we simply design the structure prior loss as $\mathcal{L}_{prior} = \mathcal{L}_{1}(\mathbf{P}_r, \mathbf{P})$. For the depth optimization, we aim to maximize the similarity between $\mathbf{P}_r$ and $\mathbf{P}$ estimated by the Pearson Correlation
Coefficient (PCC)~\cite{PCC}. Besides assessing the global correlation between depths, we divide the depth into several patches with the size of $128\times128$ at each iteration, then we randomly select $50\%$ patches to calculate the average depth correlation loss as:
\begin{equation}
\begin{split}
 \mathcal{L}^{global}_{depth} &=  1 - \operatorname{PCC}(\boldsymbol{D}_r, \boldsymbol{D}) , \\  
 \mathcal{L}^{local}_{depth} &= \frac{1}{H} \sum^{H-1}_{h=0} 1 -  \operatorname{PCC}(\boldsymbol{D}^{h}_r, \boldsymbol{D}^{h}), \\
 \mathcal{L}_{depth} &= \mathcal{L}^{global}_{depth} + \mathcal{L}^{local}_{depth}.
\end{split}
\end{equation}
Therefore, the complete structural consistency loss can be expressed as:
\begin{equation}
    \mathcal{L}_{str} = 0.1\times \mathcal{L}_{prior} + 0.1\times \mathcal{L}_{depth}
\end{equation}

\paragraph{Denoising Loss} Our progressive denoising module is designed to generate a list of denoised outputs (\{$\mathbf{R}_1, \mathbf{R}_2, ..., \mathbf{R}_K, \mathbf{R} $\}) , where $K$ represents the final stage and $\mathbf{R}$ denotes the final result for current viewpoint. The denoising loss $\mathcal{L}_{de}$ deployed to ensure effective denoising:  
\begin{equation}
    \mathcal{L}_{de} =  ||\mathbf{R} - \mathbf{R}_K||^2 + \texttt{TV}(\mathbf{R}),
\end{equation}
where \texttt{TV($\cdot$)} represents the standard TV variation regularization~\cite{TV}.

\paragraph{Reconstruction Loss} Besides, in order to acquire light-invariant representation in $\mathbf{R}$, we also render the illumination component $\mathbf{L}_r$ by adding another attribute $l_i$ to Gaussians. With our rendered illumination map $\mathbf{L}_r$ and final refined output $\mathbf{R}$, we are essentially capable of reconstructing the original image for any viewpoint by element-wise multiplication. We calculate and back-propagate the reconstruction loss as following to guide the decomposition of light information and illumination-independent component: 
\begin{equation}
\begin{split}
  \mathcal{L}_{rec} = (1-\lambda)\mathcal{L}_1(&\mathbf{I}_{out}, \mathbf{I}_{in}) + \lambda\mathcal{L}_{ssim}(\mathbf{I}_{out}, \mathbf{I}_{in}),  \\
  &\mathbf{I}_{out} = \mathbf{R} \odot \mathbf{L}_r,
\end{split}
\label{eq_loss_rec}
\end{equation}
where the loss weight $\lambda$ is set to 0.2, akin to  the configuration in 3DGS~\cite{3Dgaussian}. 

Hence, the overall optimization loss is determined by:
\begin{equation}
    \mathcal{L}=  \mathcal{L}_{exp} + \mathcal{L}_{str} + \mathcal{L}_{de} + \mathcal{L}_{rec}.
\end{equation}

\begin{figure*}[t]
        \setlength{\abovecaptionskip}{1mm}
	\centering
        \begin{subfigure}[h]{1\linewidth}
		\centering
		\includegraphics[width=\linewidth]{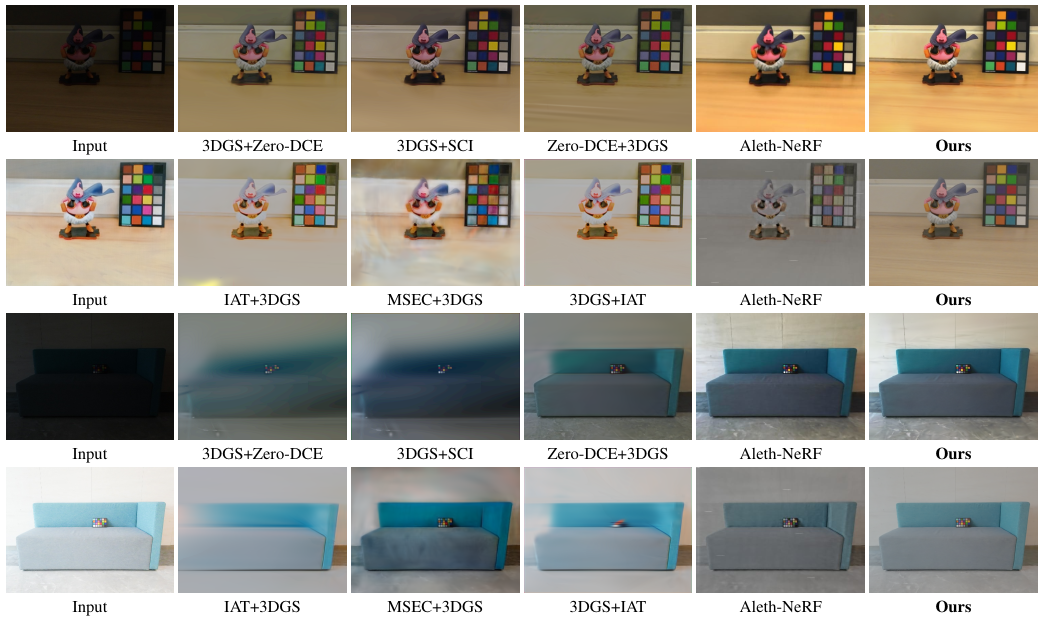}
	\end{subfigure}
    \caption{Novel view synthesis comparison in low-light and over-exposure conditions. Compared to other methods, our LITA-GS achieves more vivid enhancement results and preserves more details in novel view synthesis.}
    \label{fig_visual}
        \vspace{0mm}
\end{figure*}
\begin{table*}
\centering
\setlength{\abovecaptionskip}{2mm}
\renewcommand\arraystretch{1.3}
%\scriptsize
\setlength{\tabcolsep}{1.2pt}
\resizebox{\textwidth}{!}{
\begin{tabular}{c|ccc|ccc|ccc|ccc|ccc|ccc}
\hline
\specialrule{0.6pt}{0.5pt}{0.5pt} 
Scenes & \multicolumn{3}{c|}{``\textbf{\textit{buu}}"} & \multicolumn{3}{c|}{``\textbf{\textit{chair}}"} & \multicolumn{3}{c|}{``\textbf{\textit{sofa}}"} & \multicolumn{3}{c|}{``\textbf{\textit{bike}}"} & \multicolumn{3}{c|}{``\textbf{\textit{shrub}}"} & \multicolumn{3}{c}{\textbf{\textit{mean}}} \\ 
Method \textbar~Metrics & PSNR$\uparrow$  & SSIM$\uparrow$  & LPIPS$\downarrow$  & PSNR$\uparrow$  & SSIM$\uparrow$  & LPIPS$\downarrow$ & PSNR$\uparrow$  & SSIM$\uparrow$  & LPIPS$\downarrow$ & PSNR$\uparrow$  & SSIM$\uparrow$  & LPIPS$\downarrow$ & PSNR$\uparrow$  & SSIM$\uparrow$  & LPIPS$\downarrow$ & PSNR$\uparrow$  & SSIM$\uparrow$  & LPIPS$\downarrow$\\ \specialrule{0.6pt}{0.5pt}{0.5pt}
\hline 
NeRF & 7.12 &0.674  & 0.499  &11.05  &0.741  & 0.418 & 10.22 & 0.783 & 0.475 & 9.65 & 0.698 & 0.416 & 9.96 & 0.405 & 0.480 & 9.60 & 0.660 & 0.458 \\
Vanilla 3DGS &7.27  &0.690   &0.531  &11.13  &0.784  &0.371 &10.01  &0.766  &0.488  &9.56  &0.711  &0.392  &10.33  &0.642  &0.298  &9.66  &0.719  &0.416\\ \hline
%\large{NeRF}                    & \enspace 7.12/0.674/0.499  & 11.05/ 0.741/ \textbf{0.418}  & 10.22/ 0.783/ 0.475  & \enspace 9.65/ 0.698/ 0.416  & \enspace 9.96/ 0.405/ \underline{0.480}  & \enspace 9.60/ 0.660/ \underline{0.457}
\multicolumn{19}{c}{NeRF / 3DGS + Exposure Correction Methods} \\ \hline
NeRF + IAT & 14.11 & 0.780 & 0.433& 19.24 & 0.810 &0.491& 16.60 & 0.837 & 0.459 & 17.73 & 0.760 & 0.394 & 14.05 & 0.381 & 0.499 & 16.35 & 0.714 &0.455\\
NeRF + MSEC & 16.13 & 0.800 &0.427& 15.60 & 0.786 & 0.472 &16.56  & 0.807 & 0.495 & 12.60 & 0.716 &0.465  & 13.66 & 0.332 & 0.509 & 14.91 & 0.688 & 0.474\\ 
3DGS + IAT &16.38  &0.831  &\colorbox{orange!30}{0.344}   &18.62  &\colorbox{red!26}{0.865}  &\colorbox{orange!30}{0.262}  &18.40  &0.846  &0.383  &\colorbox{yellow!36}{20.28}  &0.753  &\colorbox{yellow!36}{0.336}  &16.47  &\colorbox{orange!30}{0.762}  &\colorbox{red!26}{0.190}  &\colorbox{yellow!36}{18.03}  &\colorbox{orange!30}{0.811}  &\colorbox{orange!30}{0.303}\\
3DGS + MSEC &\colorbox{orange!30}{16.78}  &\colorbox{orange!30}{0.852}  &\colorbox{yellow!36}{0.348}  &\colorbox{yellow!36}{19.44}  &\colorbox{orange!30}{0.858}  &\colorbox{yellow!36}{0.266}  &19.11  &0.806  &\colorbox{orange!30}{0.357}  &17.23  &\colorbox{yellow!36}{0.789} &0.359  &\colorbox{orange!30}{17.11}  &\colorbox{yellow!36}{0.748}  &0.243  & 17.93 & \colorbox{orange!30}{0.811} & \colorbox{yellow!36}{0.315}\\   \hline

\multicolumn{19}{c}{Exposure Correction Methods + NeRF / 3DGS} \\ \hline
IAT + NeRF & 16.22 & 0.815 & 0.486& 18.98 & 0.799 & 0.503 & 18.45 &\colorbox{yellow!36}{0.849}  & 0.478 & 19.63 & 0.776 & 0.408 & 15.63 & 0.434 & 0.477 & 17.78 & 0.735 & 0.470\\ 
MSEC + NeRF & 15.53 & 0.817 & 0.499 & 16.95 & 0.758 & 0.580 & \colorbox{orange!30}{19.60} & 0.817 & 0.498 & 18.90 & 0.725 & 0.483 & 15.48 & 0.400 &0.499 & 17.29& 0.703 &0.512 \\   

IAT + 3DGS &16.49  &0.834  &0.351   &18.50  &0.822  &0.394  &17.07  &0.786  &0.503  &\colorbox{orange!30}{20.38}  &\colorbox{red!26}{0.806}  &\colorbox{orange!30}{0.318}  &\colorbox{yellow!36}{16.82}  &0.691  &\colorbox{orange!30}{0.222}  &17.85  &\colorbox{yellow!36}{0.788}  &0.358\\
MSEC + 3DGS &\colorbox{yellow!36}{16.55}  &\colorbox{yellow!36}{0.843}  &0.379   &19.24  &\colorbox{yellow!36}{0.831}  &0.375  &\colorbox{yellow!36}{19.37}  &0.819  &\colorbox{yellow!36}{0.379}  &16.16  &0.698  &0.493  &16.61  &0.631  &0.288  &17.59  &0.764  &0.383  \\   \hline

\multicolumn{19}{c}{ End-to-end Methods} \\ \hline
Aleth-NeRF & \colorbox{orange!30}{16.78} & 0.805 & 0.611 & \colorbox{orange!30}{20.08} & 0.820 & 0.499 & 17.85 & \colorbox{orange!30}{0.852} & 0.458 & 19.85 & 0.773 &0.392  & 15.91 & 0.477 & 0.483 & \colorbox{orange!30}{18.09} & 0.745 & 0.489\\

\textbf{Ours} &\colorbox{red!26}{19.08}  &\colorbox{red!26}{0.885} &\colorbox{red!26}{0.288}   &\colorbox{red!26}{21.40}  &\colorbox{red!26}{0.865}  &\colorbox{red!26}{0.247}  &\colorbox{red!26}{20.01}  &\colorbox{red!26}{0.871}  &\colorbox{red!26}{0.314}  &\colorbox{red!26}{21.31}  &\colorbox{orange!30}{0.803} &\colorbox{red!26}{0.291}  &\colorbox{red!26}{19.06}  &\colorbox{red!26}{0.781}  &\colorbox{yellow!36}{0.225}  &\colorbox{red!26}{20.17}  &\colorbox{red!26}{0.841} &\colorbox{red!26}{0.273}\\  \specialrule{0.6pt}{0.5pt}{0.5pt}
\hline
\end{tabular}}
%\vspace{1pt}
\caption{We assess novel view synthesis performance in over-exposure settings by comparing generated images with ground truth
normal-light views. We report PSNR, SSIM, LPIPS and color each cell as \colorbox{red!26}{best}, \colorbox{orange!30}{second best} and \colorbox{yellow!36}{third best}. %Our method has achieved the best rendering quality, while striking a good balance between FPS and the storage memory.
}
\label{tab2}
\vspace{-4.8mm}
\end{table*}

%\cellcolor{yellow!40} third
%\cellcolor{orange!40} second
%\cellcolor{red!40} best
\section{Experiment}
\label{sec_exp}
\paragraph{Dataset} We use the LOM dataset proposed in~\cite{alethnerf} to evaluate the performance of our model in novel view synthesis. The LOM dataset comprises five real-world scenes (``buu'', ``chair", ``sofa", ``bike", ``shrub"), each containing 25 to 48 sRGB images captured by a DJI Osmo Action 3 camera under adverse lighting conditions, including low light and overexposure. For a fair comparison, we adopt the same method as AlethNeRF~\cite{alethnerf} for separating training and evaluation views. For instance, in the ``sofa" scene, we utilize the same 29 images for training and 4 images for testing.

\paragraph{Comparison Methods}
For adverse lighting conditions, we first assess the capability of vanilla NeRF and 3DGS for novel view synthesis, as detailed respectively in Tab.~\ref{tab:1} and ~\ref{tab2}. Subsequently, to evaluate the effectiveness of our proposed end-to-end approach in simultaneously rendering novel views and correcting lighting, we employ lightweight advanced image enhancement techniques (Zero-DCE~\cite{Zero-DCE}, SCI~\cite{SCI}) and exposure correction methods (IAT~\cite{IAT}, MSEC~\cite{MSEC}) as pre- and post-processing steps for vanilla NeRF~\cite{nerf} and 3DGS~\cite{3Dgaussian}. Finally, we compare with Aleth-NeRF~\cite{alethnerf}, a current end-to-end method capable of directly rendering on sRGB images.

\paragraph{Implementation Details} For each 3D Gaussian primitive, the dimensions for the introduced attributes (illumination-invariant structure, illumination feature, depth, and noise representation) is set to 1, 3, 1, 3, respectively. Our implementation leverages the PyTorch framework, adapting the CUDA kernel for rasterization to render the structure prior, depth map, illumination component, and noise representation. We utilize COLMAP to initialize the 3D Gaussian positions and estimate camera poses. Starting with a spherical harmonics degree of one, we increment the degree by one every 1,000 iterations until reaching the maximum degree of three. LITA-GS is optimized over 15,000 iterations per scene, employing the adaptive density control from 3DGS to densify and prune the Gaussian primitives during the first 5,000 iterations. Our LITA-GS demonstrates rapid convergence, completing training within 15 minutes in one RTX NVIDIA 3090. 

\paragraph{Performance Comparison} We compare the performance of our LITA-GS with current SOTA methods on low-light scenes, and we report the quantitative results as Tab.~\ref{tab:1}. Specifically, the baseline models (NeRF and 3DGS), trained in a supervised manner to reconstruct the scenes at their original exposure levels, exhibit significantly low similarity with ground truth images captured under normal lighting conditions. Moreover, despite employing image enhancement techniques on the rendered outputs of NeRF and 3DGS, the quantitative results remain unsatisfactory. When these image enhancement methods are employed as pre-processing tools and the baseline models are optimized using the enhanced images, we observe improved results in certain scenes, such as ``SCI+3DGS'' on the ``chair'' scene. Nonetheless, these image enhancement methods exhibit significant variability in performance across different scenes, leading to considerable variations in the final rendering results of NeRF and 3DGS across scenes. Therefore, leveraging image enhancement methods as pre- or post-processing tools lacks stability and reliability. 

In contrast, end-to-end methods are capable of achieving more stable results, and our LITA-GS surpasses current SOTA performance by 1.27 dB in PSNR, 0.075 in SSIM. As presented in Tab.~\ref{tab2}, we observe similar performance pattern for over-exposed scenarios and our proposed LITA-GS achieves the excellent results.
Fig.~\ref{fig_visual} shows qualitative visualization results in low light and over-exposure settings, with the comparison of current SOTA methods, we can see that our method can achieve more vivid enhancement results while also preserving more details in novel view synthesis. Other methods, however, result in significant loss of detail and inadequate brightness adjustments.

\paragraph{Ablation Study}
\label{sec_ablations}
\begin{table}[t]
\centering
\renewcommand\arraystretch{1.1}
\setlength{\tabcolsep}{3pt}
\begin{tabular}{c ccc}
\toprule
\textbf{Configuration} & PSNR$\uparrow$ & SSIM$\uparrow$ & LPIPS$\downarrow$ \\
\midrule
w/o PDM  &22.36 & 0.810 & 0.289 \\
w/o depth $\mathbf{D}_r$ & 22.53 & 0.799 & 0.295 \\
w/o illumination-invariant $\mathbf{P}_r$ & 22.19 & 0.782 & 0.327 \\
w/o PDM \& $\mathbf{D}_r$ \& $\mathbf{P}_r$ & 21.55 & 0.764& 0.359 \\
Full LITA-GS & \textbf{22.75} & \textbf{0.819} & \textbf{0.282}\\
\bottomrule
\end{tabular}
\caption{Ablation results on the ``bike'' with low-light condition. Our full LITA-GS achieves the best performance in terms of all metrics and removing any component form LITA-GS leads to obvious performance drop, highlighting the rationality of the design of our LITA-GS.}
\label{tab_abla}
\end{table}

To verify the effectiveness of our proposed LITA-GS, we conduct extensive experiments with four configurations of our method: 1) removing the progressive denoising module (w/o PDM), 2) removing the rendering of the depth map (w/o $\mathbf{D}_r$), 3) removing the rendering of the illumination-invariant structure prior (w/o $\mathbf{P}_r$), 4) removing PDM, $\mathbf{D}_r$, and $\mathbf{P}_r$ simultaneously (w/o PDM \& $\mathbf{D}_r$ \& $\mathbf{P}_r$). The quantitative results are reported in Tab.~\ref{tab_abla}. 

When the progressive denoising module is removed, the scene’s geometric structure is well-learned, as evidenced by its satisfactory SSIM; however, the original image's inherent noise remains inadequately suppressed, leading to a marked decrease in PSNR (-0.39 dB). Conversely, excluding the rendering of depth or the illumination-invariant structure from the lighting-agnostic spatial structure rendering mechanism results in a pronounced drop in SSIM, underscoring their critical role in accurately reconstructing the scene’s structure and spatial geometry. Furthermore, when both the progressive denoising module and the rendering of $\mathbf{D}_r$ \& $\mathbf{P}_r$ are excluded, the remaining framework is solely optimized using Eq.~\ref{eq_loss_enh} and Eq.~\ref{eq_loss_rec}. Although its performance falls significantly short of the full LITA-GS and the aforementioned setups, it still surpasses other methods listed in Tab.~\ref{tab:1}, thereby validating the effectiveness of our adopted loss function for optimization.

\section{Conclusion}
\label{sec_conclusion}
We present \textbf{LITA-GS}, a novel illumination-agnostic view synthesis approach that leverages reference-free 3DGS and physical priors. First, given the challenges of SfM estimation in representing scene structure and details under adverse lighting, we establish a physical prior extraction pipeline to robustly capture structural information from images with low illumination or high exposure. Secondly, we develop lighting-agnostic structure rendering process, which integrates the extracted structure prior for guidance. Furthermore, we employ a progressive denoising module for noise suppression. Extensive experiments demonstrate that LITA-GS outperforms current SOTA methods, achieving faster convergence and rendering speed.

\clearpage
\newpage
{
    \small
    \bibliographystyle{ieeenat_fullname}
    \bibliography{main}
}

\end{document}